\documentclass{article}
\usepackage{spconf,amsmath,graphicx}
\usepackage{xspace}
\usepackage{booktabs}

\usepackage{amsmath,amssymb}
\usepackage{multirow}
\usepackage{enumitem}
\usepackage{soul}
\setuldepth{Berlin}

\makeatletter
\DeclareRobustCommand\onedot{\futurelet\@let@token\@onedot}
\def\@onedot{\ifx\@let@token.\else.\null\fi\xspace}

\def\eg{\emph{e.g}\onedot} 
\def\ie{\emph{i.e}\onedot} 
 
\def\etc{\emph{etc}\onedot}

\makeatother

\usepackage[colorlinks,
            linkcolor=blue, 
            anchorcolor=blue,  
            citecolor=blue,
            ]{hyperref}
\title{A Unified Framework for Masked and Mask-Free Face Recognition via Feature Rectification}
\name{Shaozhe Hao$^{1}$ \qquad Chaofeng Chen$^2$ \qquad Zhenfang Chen$^3$ \qquad Kwan-Yee K. Wong$^{1}$}
  
\address{$^{1}$The University of Hong Kong \\
  $^{2}$Nanyang Technological University \\
  $^3$MIT-IBM Watson AI Lab}
\begin{document}
%
\maketitle
\begin{abstract}
Face recognition under ideal conditions is now considered a well-solved problem with advances in deep learning. Recognizing faces under occlusion, however, still remains a challenge. Existing techniques often fail to recognize faces with both the mouth and nose covered by a mask, which is now very common under the COVID-19 pandemic. Common approaches to tackle this problem include 1) discarding information from the masked regions during recognition and 2) restoring the masked regions before recognition. Very few works considered the consistency between features extracted from masked faces and from their mask-free counterparts. This resulted in models trained for recognizing masked faces often showing degraded performance on mask-free faces. In this paper, we propose a unified framework, named Face Feature Rectification Network ({\em FFR-Net}), for recognizing both masked and mask-free faces alike. We introduce rectification blocks to rectify features extracted by a state-of-the-art recognition model, in both spatial and channel dimensions, to minimize the distance between a masked face and its mask-free counterpart in the rectified feature space. Experiments show that our unified framework can learn a rectified feature space for recognizing both masked and mask-free faces effectively, achieving state-of-the-art results. Project code: \textcolor{magenta}{https://github.com/haoosz/FFR-Net}
\end{abstract}
\section{Introduction}
\label{sec:intro}
Face recognition has practical significance in our daily life. Recent advances of deep neural networks have resulted in face recognition systems~\cite{taigman2014deepface,sun2014deepid1,sun2014deepid2} that achieve over 99\% recognition accuracy under ideal conditions~\cite{LFWTech,kemelmacher2016megaface}. However, in real-world scenarios, face recognition systems often suffer from performance degradation due to occlusion by sunglasses, masks, \etc. In particular, masked face recognition has recently been a hot topic under the COVID-19 pandemic.

To recognize occluded faces, some researchers~\cite{lingxue2019pdsn,wan2017occlusion} proposed to discard features from the occluded regions. Although this approach can avoid distraction caused by the corrupted features of the occluded regions, it also results in loss of global face information which might be inferred from the discarded regions. Others~\cite{cheng2015robust,zhao2018robust} proposed to restore the occluded face regions before recognition. They commonly used mean squared errors in the pixel space to train generative adversarial networks~\cite{goodfellow2014generative} to restore a face. Despite the fact that the restored face may be visually appealing, they fail to preserve identity information in the feature space. In fact, very few works considered the consistency between features extracted from occluded faces and from their occlusion-free counterparts. In this case, models trained for recognizing occluded faces often show degraded performance on occlusion-free faces.


In this paper, we specifically focus on masked face recognition and propose a unified framework for masked and mask-free face recognition. In our model, the corrupted features from masked faces can be mapped to a rectified feature space with better identity separability. We adopt SENet~\cite{hu2018squeeze} trained on mask-free faces as our feature extractor. We introduce rectification blocks (RecBlocks) on top of extracted features. RecBlocks are trained to maximize the consistency between masked faces and their mask-free counterparts in the rectified feature space. Instead of directly transforming the extracted features, RecBlocks output a linear transformation matrix ${\bf M}$ applied to the extracted features. 
Ideally, ${\bf M}$ should be an identity matrix for mask-free faces and should suppress features from the masked regions for masked faces. We carry out rectifications in both spatial and channel dimensions using two parallel RecBlocks. In addition to the commonly used classification loss, we employ an identity loss and a triplet loss to train our rectification network end-to-end. In summary, our main contributions are:
\begin{itemize}[itemsep=-3pt, leftmargin=20pt, topsep=0pt]
    \item We introduce a unified framework, named Face Feature Rectification Network ({\em FFR-Net}), for masked and mask-free face recognition. Experiments show that our framework achieves the best average performance on mixed datasets (mask and mask-free faces) with a single model. 
    \item We propose rectification blocks (RecBlocks) to rectify features of masked or mask-free faces in both spatial and channel dimensions. RecBlocks can maximize the consistency between masked faces and their mask-free counterparts in the rectified feature space.
\end{itemize}

\section{Methodology}

\begin{figure}[t]
    \centering
    \includegraphics[width=0.99\linewidth]{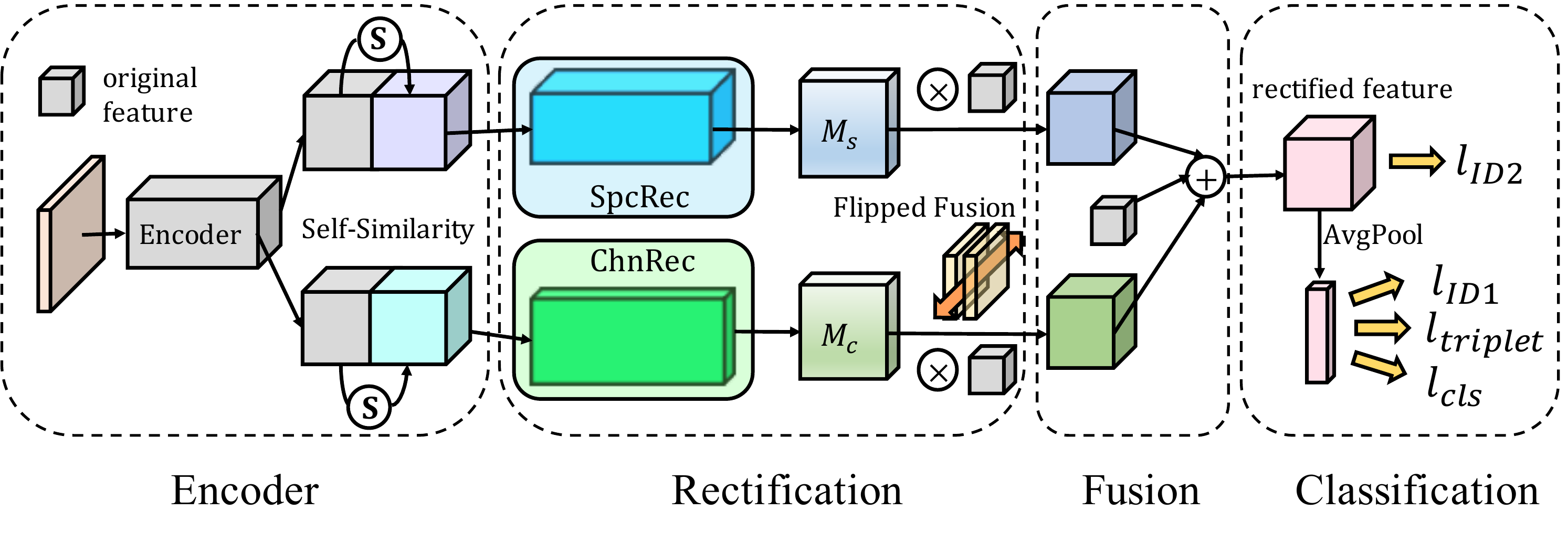}
    \caption{Overview of FFR-Net. We first extract features of masked/mask-free faces with a pre-trained encoder. Then, we use channel and spatial rectification blocks to rectify features to a unified space. Finally, the original and rectified features are fused and sent to the classifier. \textcircled{\small{S}} represents self-similarity, \textcircled{\small{$\times$}} represents matrix multiplication and \textcircled{\small{+}} represents concatenation and fusion.}
    \label{fig:overview}
\end{figure}

In this section, we will introduce the proposed feature rectification block and describe our training objectives in detail. Fig.~\ref{fig:overview} gives an overview of Face Feature Rectification Network (\textit{FFR-Net}). 

\subsection{Channel-and-Space Rectification}

Consider a training set $\{x_i^0, x_i^1\}_{i=1}^N$ with $N$ pairs of mask-free faces $x_i^0$ and masked faces $x_i^1$. Let $F$ denote the face encoder that extracts a feature map $f_i^j$ from $x_i^j$, \ie, $f_i^j = F(x_i^j)$, where $j \in \{0, 1\}$. $R$ denotes a rectification module that maps $f_i^j$ to $\hat{f}_i^j$ in the rectified feature space, \ie, $\hat{f}_i^j = R(f_i^j)$. Our goal is to train a rectification module to perform the mapping $f_i^j \rightarrow \hat{f}_i^j$ so as to maximize the recognition accuracy for both $x_i^0$ and $x_i^1$ based on the rectified features $\hat{f}_i^j$. For simplicity, we omitted the subscripts and superscripts $i, j$ in this section.
 
To fully exploit the inner structure of encoded features, we conduct the rectification process in both channel dimension and spatial dimension,~\ie, \textit{ChnRec} and \textit{SpcRec}. The main idea is to output a linear mapping of original encoded features $f \in \mathbb{R}^{C \times (\!H \!\times\! W\!)}$, where $C$ and $H\!\times\! W$ denote channel and spatial size respectively. The rectification can be expressed as   
\begin{gather}
     \label{eq:rec} {\hat{f_c}} = M_c \otimes f \qquad {\hat{f_s}} = f \otimes M_s
\end{gather}
where $\otimes$ is matrix multiplication, $M_c \in \mathbb{R}^{C \times C}$ and $M_s \in \mathbb{R}^{(\!H\!\times \!W\!) \times (\!H\!\times\! W\!)}$ are channel and spatial rectification matrices. $M_c$ and $M_s$ are generated with \textit{ChnRec} and \textit{SpcRec}. The structures are described in Sec.~\ref{sec:implement}. The final rectified feature $\hat{f}$ is obtained by fusing $\hat{f_c}$, $\hat{f_s}$ with the original feature $f$. 

Besides, We also discover two important properties that are helpful to predict $M$: self-similarity and spatial symmetry.

\textit{1. Self-similarity.} Aligned mask-free faces usually have similar facial structure, while this pattern can be corrupted due to occurrence of masks. Inspired by this observation, we introduce the self-similarity feature as extra input for \textit{ChnRec} and \textit{SpcRec}. Specifically, the self-similarity features are defined as the cosine similarity matrices between feature vectors in both spatial and channel dimension which can be formulated as below  
\begin{align}
   \label{eq:ss_spc} S_{i,j}^s &= \langle\frac{f_i}{\|f_i\|_2},\frac{f_j}{\|f_j\|_2}\rangle, i,j \in \{0,1, \cdots, H \!\times \!W\!-\!1 \} \\
   \label{eq:ss_chn} S_{m,n}^c &= \langle\frac{f_m}{\|f_m\|_2},\frac{f_n}{\|f_n\|_2}\rangle, m,n \in \{0,1, \cdots, C\!-\!1 \} 
\end{align}
where $\langle\cdot,\cdot\rangle$ is inner product, $f_i \in \mathbb{R}^C$ is the $i$-th feature vector in spatial dimension, and $f_m \in \mathbb{R}^{H\times W}$ is the $m$-th feature vector in channel dimension. $S^s \in \mathbb{R}^{(\!H \!\times\! W\!) \times (\!H\! \times\! W\!)}$ and $S^c \in \mathbb{R}^{C \times C}$ are spatial and channel self-similarity matrices respectively. We concatenate $S^s$ and $S^c$ with original feature $f$ to obtain $f^s \in \mathbb{R}^{(C+(\!H\! \times\! W\!)) \times \!H\! \times\! W\!}$ and $f^c \in \mathbb{R}^{C \times ((\!H \!\times\! W\!)+C)}$ as inputs for \textit{ChnRec} and \textit{SpcRec} respectively. 

\textit{2. Spatial Symmetry.}
Channel rectification is aimed to select those channels which extract features from mask-free regions and discard those from masked parts. However, the selected channels might only contain information from single side of the face, \eg the left eye or the right cheek. Because $f^c$ ignores spatial information, channel rectification may not be able to preserve the spatial symmetry property of mask-free faces. Thus, we introduce flipped fusion block as the last step of \textit{ChnRec}. The flipped fusion block merges the original feature with the horizontally flipped feature, which ensures outputs from \textit{ChnRec} are still spatially symmetric.

\subsection{Loss Definition}
\label{sec:loss}
The rectified feature space is supposed to have three properties: 1) rectified features of the same identity, no matter masked or not, should give similar representations to the original mask-free one. 2) masked features should give a rectification tendency to approach original mask-free encoded features. 3) each pair of rectified features should produce the best classification result for the masked face recognition task. Accordingly, we leverage three losses for identity consistency, mask robustness and accurate classification.

\noindent{\textbf{Identity consistency.}} We propose 2 components in identity loss, i.e., identical facial representations ($\mathcal{L}_{ID1}$) and identical self-similarity patterns ($\mathcal{L}_{ID2}$). Identity loss aims to achieve: (1) rectify the masked feature and (2) keep the mask-free feature uncorrupted. Therefore, each component has 2 parts to simultaneously penalize masked and mask-free data. The formulation is as follows:
\begin{align}
    \mathcal{L}_{ID1} = \frac{1}{N}\sum\limits_i \| \hat{f}_i^1 - f_i^0 \|_2 + \| \hat{f}_i^0 - f_i^0 \|_2 \\
    \mathcal{L}_{ID2} = \frac{1}{N}\sum\limits_i \| \hat{S}_i^1 - S_i^0 \|_2 + \| \hat{S}_i^0 - S_i^0 \|_2 
\end{align}
where $f_i$ and $S_i$ are the $i$-{th} encoded feature and its self-similarity matrix, and the superscript $\{0,1\}$ represents the mask-free and masked data respectively. 

\noindent{\textbf{Mask robustness.}}
A variant triplet loss is adopted. Specifically, for a masked face sample, triplet loss minimizes the distance between its rectified representation (the anchor $f_i^a$) and the original representation of its mask-free counterpart (the positive $f_i^p$) while maximizing the distance between the rectified representation (the anchor $f_i^a$) and the original representation of itself (the negative $f_i^n$). Triplet loss $\mathcal{L}_{triplet}$ with cosine distance is formulated as:
\begin{gather}
p_i = 1 - cos(f_i^a, f_i^p)	\qquad	n_i = 1 - cos(f_i^a, f_i^n)	\\
 \mathcal{L}_{triplet} = \frac{1}{N} \sum\limits_i [(p_i - n_i) + m]_+ 
\end{gather}
where $p_i$ and $n_i$ represent $i$-{th} positive and negative item of triplet loss respectively. The margin value $m$ is set to $0.1$. 

\noindent{\textbf{Accurate classification.}} We choose CosFace~\cite{wang2018cosface} as our classification loss, formulated as below 
\begin{equation}
\mathcal{L}_{cls}=\frac{1}{N}\sum^N_{i=1}{\!-\!\log{\frac{e^{s(\cos{\theta_{y_i,i}}-m)}}{e^{s(\cos{\theta_{y_i,i}}-m)}\!+\!\sum^n_{j=1, j\neq{y_i}}{e^{s\cos{\theta_{j,i}}}}}}}
\end{equation}
where $s$ is set to $30.0$ and $m$ is set to $0.4$.
We notice that ArcFace~\cite{deng2018arcface} demonstrates better mask-free face classification performance, but empirically CosFace is better in our framework. More experimental details are in Sec.~\ref{sec:ablation}. 

\section{Implementation}
\label{sec:implement}

\noindent{\textbf{Network Structure.}}
We use SOTA face recognition model SENet50 as face encoder~\cite{deng2018arcface}. We employ \{Conv-PReLU-BN\} residual convolutional structure for \textit{SpcRec} and \{Linear-PReLU-Linear\} fully connected structure for \textit{ChnRec}. Both have 9 layers with Sigmoid at last. For feature fusion, we simply concatenate three features and feed into a \{Conv-PReLU-BN\} block.

\noindent{\textbf{Model Enhancement.}}
Considering our model might sacrifice the accuracy of clean face recognition slightly, we propose a model enhancement method (\textit{FFR-Net+}). For each pair of faces, we combine the similarity score $\hat{s}$ from FFR-Net and the score $s$ from raw SENet50. Note that SENet is a part of our model so no extra computational cost will be added. The new score $s_n$ is formulated as $s_n = max(s, \hat{s}) + \hat{s}$.

\noindent{\textbf{Training Details.}}
The encoder SENet50 pre-trained on refined MS1M~\cite{guo2016ms} is fixed and only FFR-Net is trainable. We train models on CASIA-WebFace~\cite{yi2014learning} and the simulated masked data. We use Adam~\cite{kingma2015adam} with $lr=0.01$ as optimizer. 

\noindent{\textbf{Real-world Data.}}
The collected real-world masked face dataset is refined from RMFD proposed by \cite{wang2020masked}, consisting of 100 identities, 512 images and 782 pairs of a masked face and a clean face. Evaluation on real data is shown in Sec.~\ref{sec:performance}.

\noindent{\textbf{Mask Simulation.}}
We choose MaskTheFace~\cite{anwar2020masked} repository\footnote{\textcolor{magenta}{https://github.com/aqeelanwar/MaskTheFace}} as our mask generator. The type of masks to simulate single face is randomly chosen for robustness.

\section{Experimental Results}
\subsection{Evaluation Metric}
\label{sec:metric}
\noindent{\textbf{Average Accuracy.}}
Performance imbalance exists between mask-free and masked faces in many face recognition models. Therefore, datasets of all mask-free or all masked faces are not sufficient for fair comparison. We introduce a series of testing datasets $\{D_i|i=0.0, 0.1, \ldots, 1.0\}$ in which $i$ is randomly masked faces ratio. Accuracy is obtained on all datasets respectively. The mean accuracy is used as the key evaluation metric, denoted as Avg. in following experiments. Besides, we also consider another 2 criterion for comparison: 1) Mask-free: accuracy only on mask-free data and 2) Masked: accuracy only on masked data.

\noindent{\textbf{Baseline Models.}}
We consider four baseline models. 1) \textit{SENet50}: the face recognition model, i.e., our encoder. 2) \textit{f-SENet50}: the same network structure as the first one but finetuned by the simulated CASIA-WebFace dataset. 3) \textit{SENet50-convs}: the backbone model SENet50 with additional convolutional layers. 4) \textit{PDSN}: pairwise differential siamese network (PDSN)~\cite{lingxue2019pdsn}, the SOTA method based on feature discarding mask to eliminate harmful interference of corrupted features.

\subsection{Performance}
\label{sec:performance}
\begin{figure}[t]
\begin{minipage}[!t]{0.49\linewidth}
  \centering   
   \includegraphics[width=.99\linewidth]{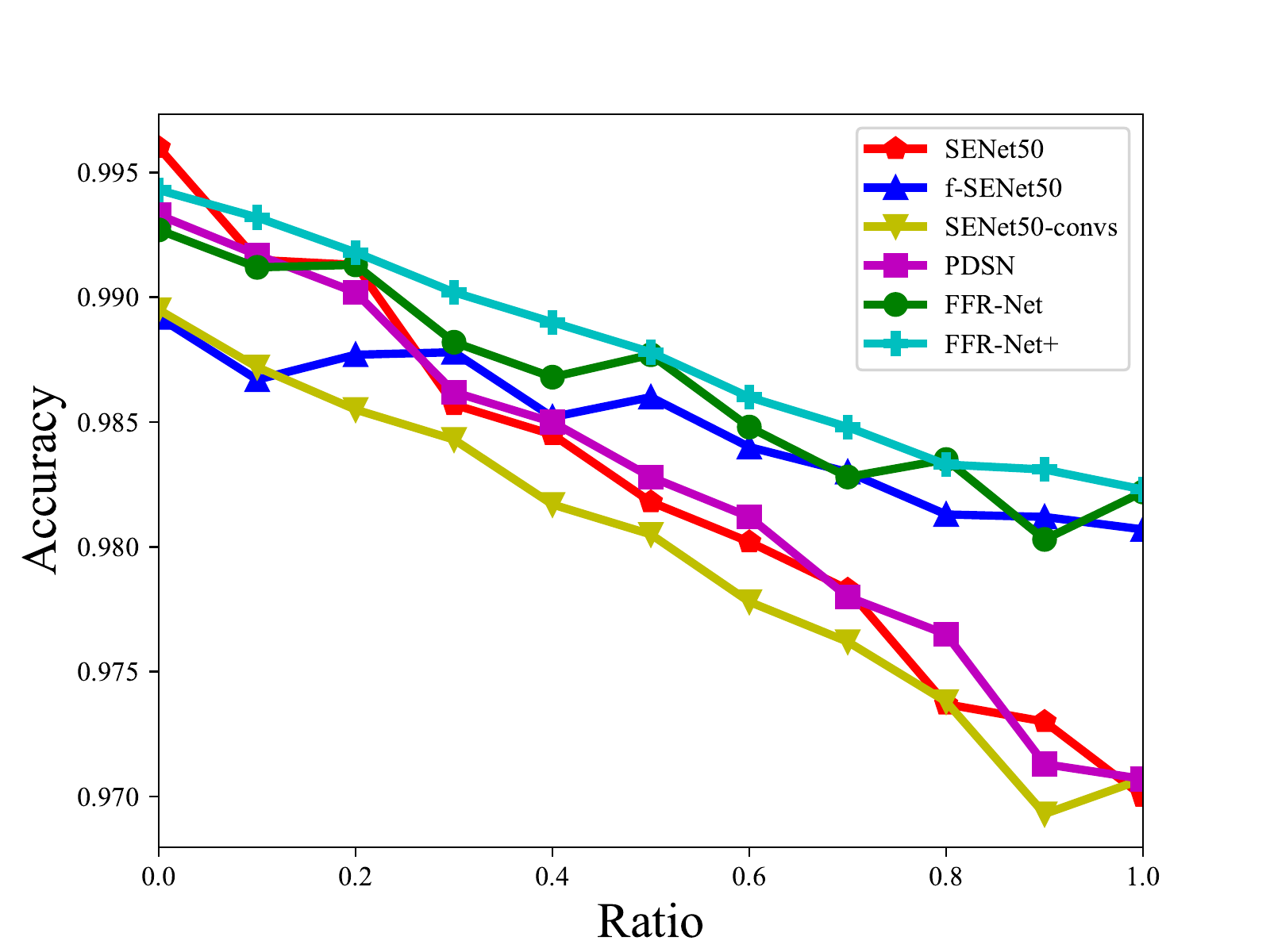}    
\caption{Acc@Ratio on LFW.} 
\label{fig:acc-lfw} 
  \end{minipage}
\begin{minipage}[!t]{0.49\linewidth}
  \centering   
   \includegraphics[width=.99\linewidth]{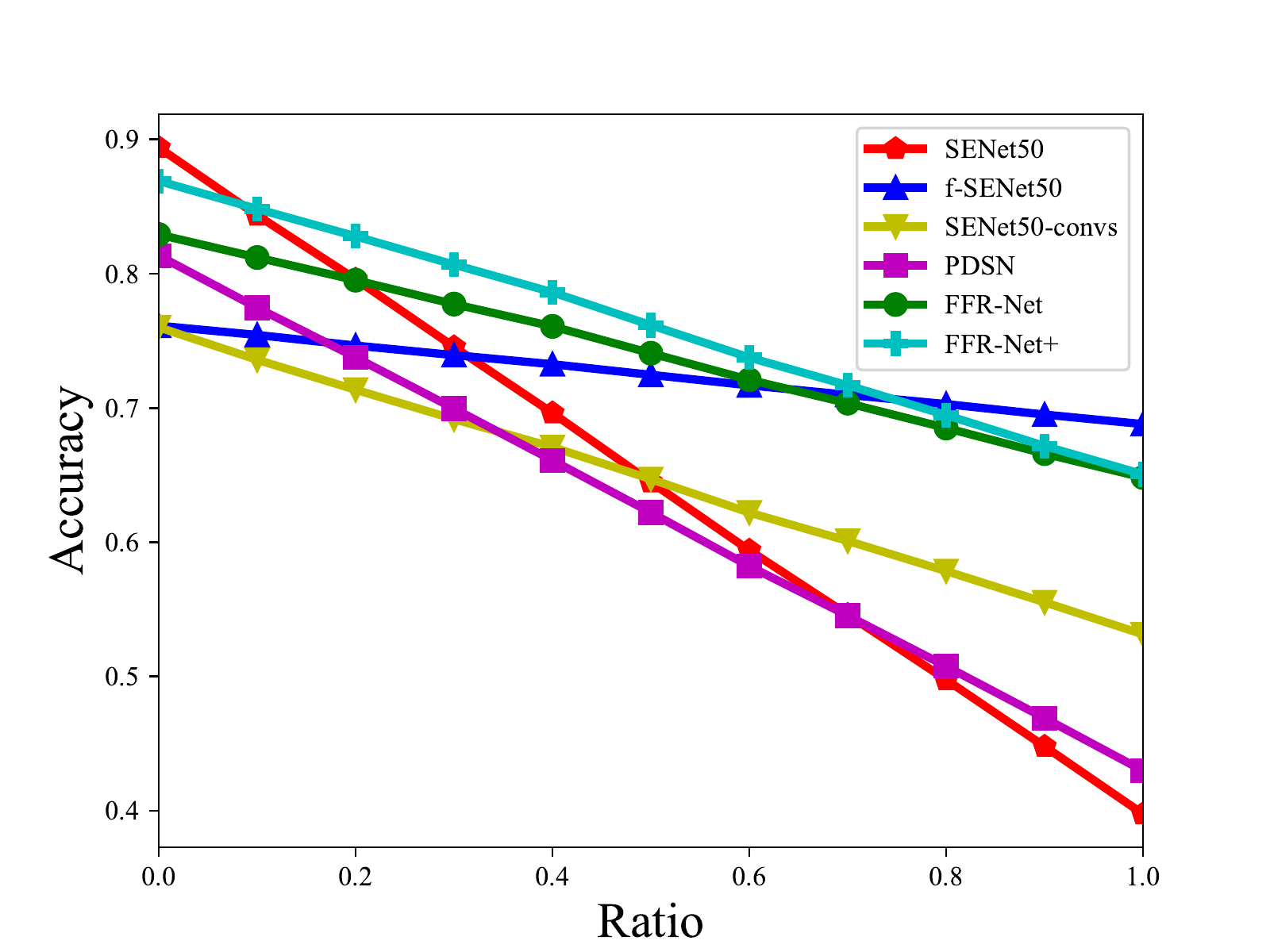}    
\caption{Acc@Ratio on MF1.} 
\label{fig:acc-megaface} 
  \end{minipage}
\end{figure}

\noindent{\textbf{Face Verification.}}
We conduct evaluation on LFW~\cite{LFWTech} and follow the standard protocol. We report the verification accuracy in Fig.~\ref{fig:acc-lfw} and Table~\ref{tab:lfw}. We can observe \textit{f-SENet50} improves with masked faces by $1.07\%$ but decreases by $0.68\%$ when mask free, because the finetuned model corrupts the original feature space to some extent. \textit{SENet50-convs} shows no improvement, indicating only increasing convolutional layers is not sufficient. Our model increases the average accuracy by $0.41\%$ relative to \textit{SENet50}, outperforming the others by $+0.16\%$. \textit{FFR-Net+} further improves $0.14\%$. On real-world masked data, our method improves \textit{SENet50} by $3.08\%$. \textit{FFR-Net} slightly outperforms \textit{FFR-Net+} because the enhancement method is to improve the performance with mask-free data instead of masked conditions.
\vspace{-7pt}
\begin{table}[h]
  \caption{Face verification accuracy on LFW and real data}
    \label{tab:lfw}
    \centering
    \resizebox{.45\textwidth}{!}{
    \begin{tabular}{c|c|c|c|c|c}
    \toprule
       Method & \#Image & Mask-free & Masked & Avg. & Real data \\ \midrule
       
       CosFace & 5M & 99.73\% & ---  & --- & ---\\
       ArcFace & 0.49M & 99.53\% & --- & --- & ---\\\bottomrule\toprule
       \textit{SENet50} & 5M & \textbf{99.60\%} & 97.00\% & 98.23\% & 90.00\% \\ 
       \textit{f-SENet50} & 0.98M & 98.92\% & 98.07\% & 98.48\% & 89.36\% \\ 
       \textit{SENet50-convs} & 0.98M & 98.95\% & 97.07\% & 97.96\% & 86.03\% \\ 
       \textit{PDSN} & 0.98M & 99.33\% & 97.07\% & 98.25\% & 90.77\% \\ 
       \textit{FFR-Net} & 0.98M & 99.27\% & \ul{98.22\%} & \ul{98.64\%} & \textbf{93.08\%} \\
       \textit{FFR-Net+} & 0.98M & \ul{99.43\%} & \textbf{98.23\%} & \textbf{98.78\%} & \ul{92.95\%} \\ \bottomrule
    \end{tabular}
    }
\end{table}
\vspace{-7pt}

\noindent{\textbf{Face Identification.}}
We also test the model on MegaFace Challenge1~\cite{kemelmacher2016megaface} (MF1) to evaluate the performance of face identification task. The training set is viewed as small if it is less than 0.5M or viewed as large. Rank-1 accuracy with 1M distractors is used in experiments. From Fig.~\ref{fig:acc-megaface} and Table~\ref{tab:megaface}, \textit{SENet50} decreases the accuracy by $49.64\%$ due to masks. Compared to \textit{SENet50}, our method greatly improves on masked data by $25.06\%$ while keeping considerable performance on mask-free one with only $6.48\%$ accuracy dropping. Our model outperforms other methods by $+1.54\%$. \textit{FFR-Net+} further improves $2.11\%$. In addition, our model only increases FLOPS by $17.9\%$ and the number of trainable parameters is $50.6\%$ of the fixed backbone, which indicates low calculation cost of the proposed model.
\vspace{-7pt}
\begin{table}[h]
  \caption{Face identification accuracy on MF1}
    \label{tab:megaface}
    \centering
    \resizebox{.45\textwidth}{!}{
    \begin{tabular}{c|c|c|c|c|c|c}
    \toprule
       Method & protocol & FLOPS & $\#$params & Mask-free & Masked & Avg. \\ \midrule
       
       CosFace & small & 6.31G & 43.79M & 77.11\% & --- & --- \\ 
       ArcFace & small & 6.31G & 43.79M & 77.50\% & --- & --- \\ 
        CosFace & large & 6.31G & 43.79M & 82.72\% & --- & --- \\\bottomrule \toprule
       \textit{SENet50} & large & 6.31G & 43.79M & \textbf{89.40\%} & 39.76\% & 64.57\% \\ 
       \textit{f-SENet50} & large & 6.31G & 43.79M & 76.13\% & \textbf{68.80\%} & 72.47\% \\
       \textit{SENet50-convs} & large & 7.00G & 57.95M & 76.01\% & 53.14\% & 64.62\%  \\
       \textit{PDSN} & large & 6.45G & 59.05M & 81.35\% & 42.98\% & 62.21\% \\ 
       \textit{FFR-Net} & large & 7.44G & 65.94M & 82.92\% & 64.82\% & \ul{74.01\%} \\
       \textit{FFR-Net+} & large & 7.44G & 65.94M & \ul{86.91\%} & \ul{65.06\%} & \textbf{76.12\%}\\ \bottomrule
    \end{tabular}
    }
 \end{table}
 \vspace{-10pt}
 
\subsection{Ablation Study}
\label{sec:ablation}

\noindent{\textbf{Component Ablation.}}
We first test the significance of self-similarity features ($SS$), flipped fusion block ($FF$), as tag A, B in Table~\ref{tab:component-ab}. The absence of $SS$ and $FF$ respectively decreases the average accuracy. Besides, we verify the significance of three features in fusion ($\hat{f}_s$, $\hat{f}_c$ and $f$). C1 and C2 respectively exclude \textit{ChnRec} and \textit{SpcRec}, which result in $1.95\%$ and $2.30\%$ accuracy declining. This proves the efficiency of channel-and-space rectification block. C3 shows $3.26\%$ accuracy dropping without fusing the original feature. Therefore, fusing $f$ along with $\hat{f}_s$ and $\hat{f}_c$ can greatly boost the performance. 
\vspace{-7pt}
\begin{table}[h]
  \caption{Ablation study of components on MF1}
    \label{tab:component-ab}
    \centering
    \resizebox{.45\textwidth}{!}{
    \begin{tabular}{c|c|c|c|c|c|c|c|c}
    \toprule
        Tag & $SS$ & $FF$ & $\hat{f}_s$ & $\hat{f}_c$ & $f$ & Mask-free & Masked & Avg. \\ \midrule
        A & &\checkmark &\checkmark &\checkmark &\checkmark & 82.00\% & 63.29\% & 72.61\% \\ 
        B & \checkmark &  &\checkmark &\checkmark &\checkmark & 78.57\% & 56.75\% & 67.83\% \\
        C1 & \checkmark & \checkmark &\checkmark & &\checkmark & 81.69\% & 62.42\% & 72.06\% \\ 
        C2 & \checkmark & \checkmark & &\checkmark &\checkmark & 82.39\% & 61.03\% & 71.71\% \\
        C3 & \checkmark & \checkmark &\checkmark &\checkmark & & 79.06\% & 62.44\% & 70.75\% \\ \midrule
        Full & \checkmark & \checkmark &\checkmark &\checkmark &\checkmark & \textbf{82.92\%} & \textbf{64.82}\% & \textbf{74.01\%} \\ \bottomrule
    \end{tabular}
    }
\end{table}
\vspace{-7pt}

\noindent{\textbf{Loss Ablation.}}
In Table~\ref{tab:loss-ab}, we conduct ablation study on identical facial representation constraint ($\mathcal{L}_{ID1}$) and identical self-similarity pattern constraint ($\mathcal{L}_{ID2}$) as tag D. Accuracy declines by $+1.18\%$ when removing $\mathcal{L}_{ID1}$ or $\mathcal{L}_{ID2}$. Besides, we test our model with ArcFace~\cite{deng2018arcface} as tag E. ArcFace fails to train an efficient model with a $39.71\%$ drop, proving CosFace is more suitable for training of learnable module RecBlocks.
\vspace{-9pt}
\begin{table}[h]
    \caption{Ablation study of losses on MF1}
    \label{tab:loss-ab}
    \centering
    \resizebox{.45\textwidth}{!}{
    \begin{tabular}{c|c|c|c|c|c|c|c}
    \toprule
        Tag & $\mathcal{L}_{ID1}$ & $\mathcal{L}_{ID2}$ & $AF$ & $CF$ & Mask-free & Masked & Avg. \\ \midrule
        D1 & & & &\checkmark & 81.77\% & 63.10\% & 72.54\% \\ 
        D2 & \checkmark & & &\checkmark & 82.03\% & 62.95\% & 72.55\% \\
        D3 & &\checkmark & &\checkmark & 82.76\% & 62.88\% & 72.83\% \\
        E &\checkmark &\checkmark &\checkmark & & 41.45\% & 27.25\% & 34.30\% \\ \midrule
        Full &\checkmark &\checkmark & &\checkmark &  \textbf{82.92\%} & \textbf{64.82}\% & \textbf{74.01\%} \\ \bottomrule
    \end{tabular}
    }
\end{table}
\vspace{-26pt}

\subsection{Feature Space Visualization}
\begin{figure}[t]
\begin{minipage}{0.49\linewidth}
\centering
\includegraphics[scale=0.28]{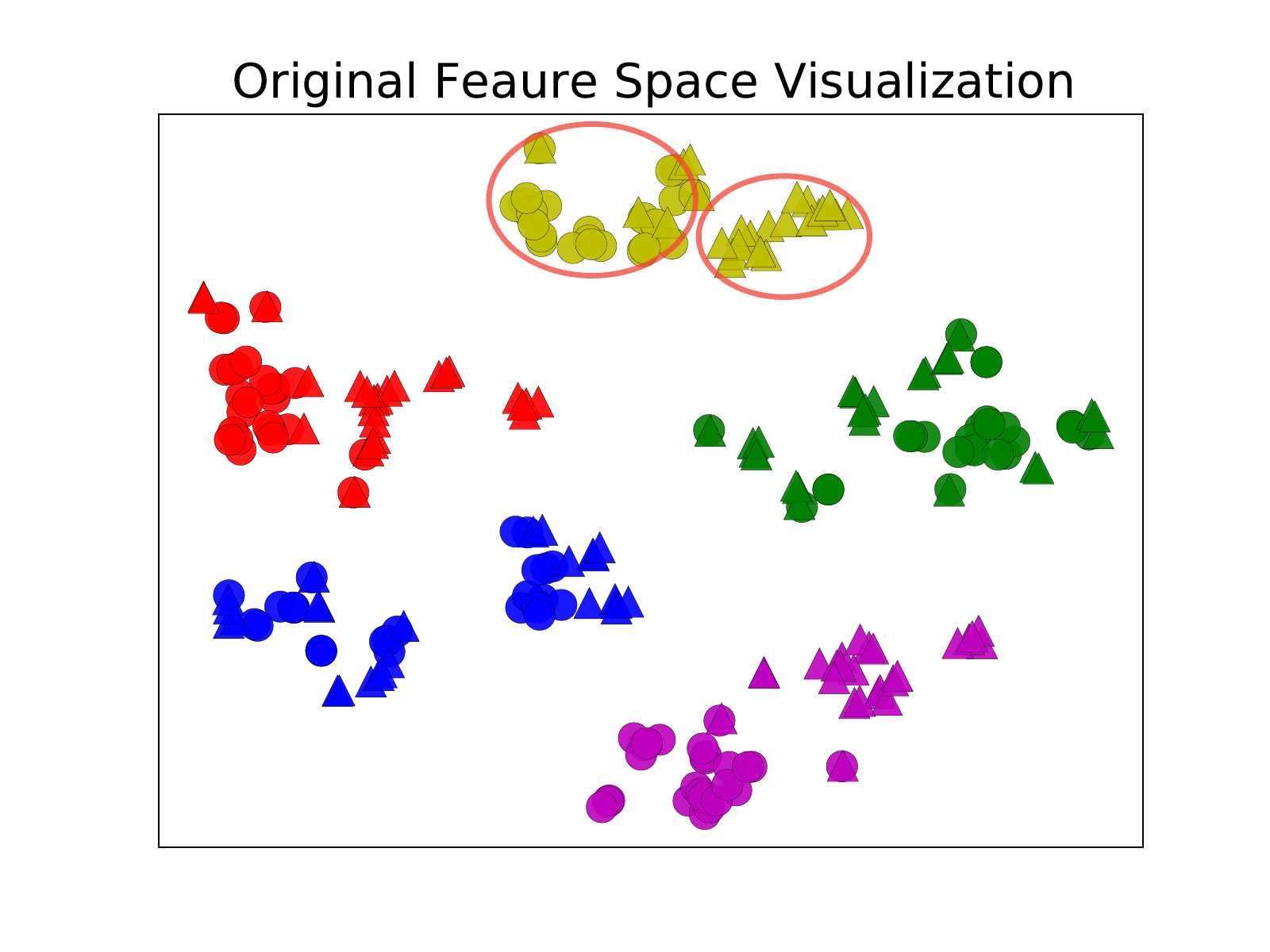} 
\end{minipage}
\begin{minipage}{0.49\linewidth}
\centering                                                          
\includegraphics[scale=0.28]{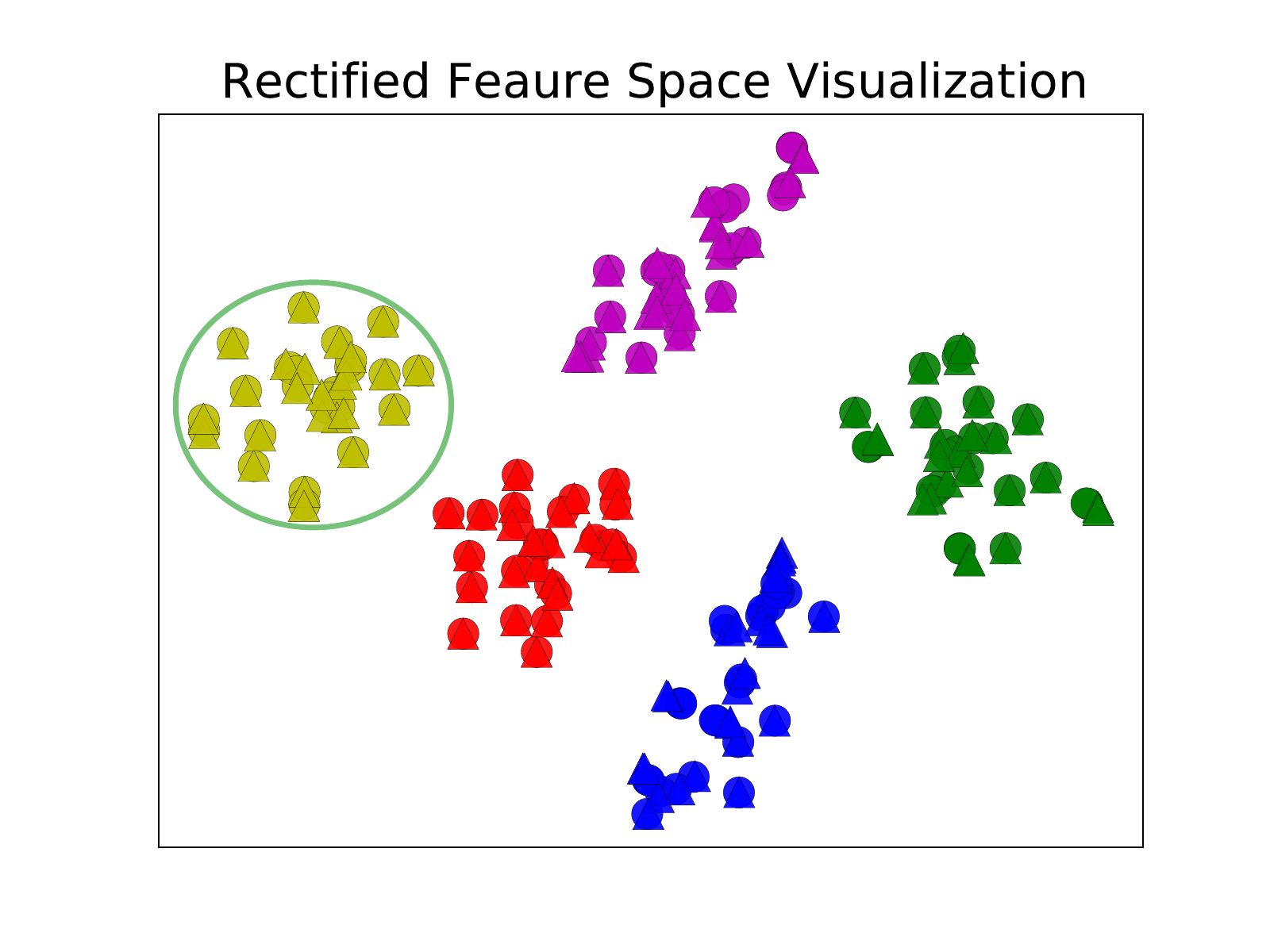}     
\end{minipage}
\caption{T-SNE visualization: colors represent different identities, $\bigcirc$ denote mask-free face and $\bigtriangleup$ denotes masked face.}                         
\label{fig:featurevis}                                                        
\end{figure}

The main idea of feature rectification is to find a proper mapping to project the original features to rectified features. We visualize the original and the rectified feature spaces using t-SNE to display the effective mapping process. The result is shown in Fig.~\ref{fig:featurevis}. In the original feature space, triangles and circles tend to gather separately and keep far away from each other. However, in the rectified feature space, triangles tend to align with circles and they are equally distributed. Compared to the original space, the rectified space possesses more compact intra-class geodesic distances and larger inter-class margins with greater similarity between the masked feature and the mask-free counterpart. This result matches our initial idea that the rectified feature space is more identity-separable and mask-resistant than the original one.

\section{CONCLUSION}
In this paper, we introduce a novel framework via feature rectification to uniformly recognize masked and mask-free faces. In this context, we propose channel-and-space rectification block to rectify the encoded facial representation at feature level in both spatial and channel dimensions and designed powerful losses to enforce ideal properties of the unified rectified feature space. As a result, our model achieves state-of-the-art performance on mixed datasets of masked and mask-free faces. Finally, we verify the contributions of the proposed blocks and losses to performance enhancement and rectification mapping.

\vfill
\pagebreak

\bibliographystyle{IEEEbib}
\bibliography{egbib}

\end{document}